# Pyramid Vector Quantization and Bit Level Sparsity in Weights for Efficient Neural Networks Inference

Vincenzo Liguori        Ocean Logic Pty Ltd        enzo@ocean-logic.com

**Abstract**
**This paper discusses three basic blocks for the inference of convolutional neural networks (CNNs). Pyramid Vector Quantization [1] (PVQ) is discussed as an effective quantizer for CNNs weights resulting in highly sparse and compressible networks. Properties of PVQ are exploited for the elimination of multipliers during inference while maintaining high performance. The result is then extended to any other quantized weights. The Tiny Yolo v3 [2] CNN is used to compare such basic blocks.**

# 1. Introduction

Neural networks (NNs) and, particularly, convolutional neural networks (CNNs) have emerged in recent years as a powerful tool for image (and other types of signals) recognition, classification and processing. The high computational cost of their inference remains a formidable obstacle in low power and low cost environment such as embedded system and IoT.

The focus of this paper is on convolutions as the most computationally intensive aspect of CNN inference. These can be reduced to dot products between a weights vector and an input vector.

We will initially discuss dot product calculations using multiply accumulators (MACs). Multipliers will be subsequently eliminated by exploiting the proprieties PVQed weights. Next this result will be improved and extended to other quantized weights. Weight compression will also be discussed. Finally, all the results will be compared using the Tiny Yolo v3 CNN as an example.

# 2. Dot Products with MACs

Lets' consider the definition of dot products between a weight vector $\vec{w}$ and an input vector $\vec{x}$ of dimensionality N:

$$\vec{w} \cdot \vec{x} = \sum_{j=0}^{N-1} w_j x_j \quad (1)$$

From the definition the use of MACs follows naturally. A MAC implementation is shown in Fig.1a. Many implementations are possible that can reach one multiply and accumulation operation per clock cycle (2 ops/cycle). Since (1) requires N multiplications and N additions, we can expect a MAC to take N cycles.

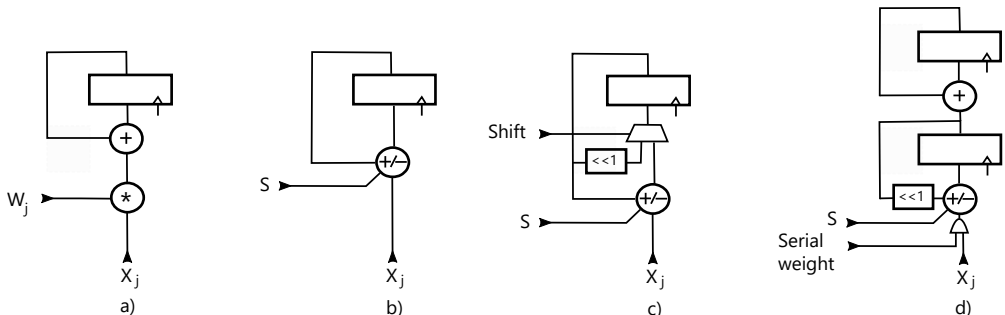

*Figure 1 Dot product hardware architectures.*

This number can be reduced if we skip the zero weights that do not make a contribution to the result. Since for many CNNs, especially after quantization, many weights are zero, this could reduce the number of cycles.

For this to work, even with many zero weights, we need an efficient mechanism to feed to the MAC only weights $w_j \neq 0$ and their indexes j to retrieve the corresponding $x_j$ input. This works well with sparse weights as it allows to jump from one non-zero weight to another, provided that we can skip all the zero ones with no clock cycles penalties.

This can be achieved by run-length encoding the weights. Each non-zero weight can be encoded as a (ZRUN,W) pair. Here ZRUN represent the number of zero weights preceding the weight W. The pair (0,0) will indicate that all the remaining weights are zero (End of Run, EOR) thus they do not contribute to the dot product and the calculation is complete.

The (ZRUN,W) pairs can also be compressed into a bitstream with an entropy encoder such as Huffman or arithmetic coding.

During the calculation of the dot product, said bitstream will be decompressed with an entropy decoder and the run-lengths expanded. This will result in a stream of weights $w_j \neq 0$ and their indexes j. The latter will be used to retrieve the corresponding $x_j$ inputs. Both $w_j$ and $x_j$ will be used by the MAC.

# 3. Pyramid Vector Quantization

This section will use the properties of Pyramid Vector Quantization[1] (PVQ), a vector quantization technique, in order to eliminate the multipliers from dot product calculations while maintaining good performances.

PVQ is defined by two integers : N, the dimensionality of the vector to be quantized and Q, the amount of quantization. A larger Q values means a better approximation of the original vector. After PVQ, a vector $\vec{w}$ will be approximated by:

$$\vec{w} \simeq \rho \hat{w} \quad (2)$$

with $\rho \geq 0, \rho \in \mathbb{R}$ and $\hat{w} \in \mathbb{Z}^N$ (all its components $\hat{w}_j$ are integers). Also :

$$\sum_{j=0}^{N-1} |\hat{w}_j| = Q \quad (3)$$

The original PVQ paper [1] claims that PVQ is superior to scalar quantization for vectors whose components have Laplacian or Gaussian distributions. It has been shown in [3] that in many CNNs the distribution of the weights is approximately Laplacian/ Gaussian. The author has verified this in [4] and also that applying PVQ to such weights results only in a small loss of accuracy compared to the original CNN for 1 < Q/N < 2. The first layer requires less quantization with Q/N=~3-4 while fully connected layers can be quantized more, with Q/N=~1/3.

The author has shown in [5] the dot product between any vector and a PVQed vectors requires only one multiplication and Q-1 additions. In fact, for any vector $\vec{x}$, the dot product $\vec{w} \cdot \vec{x}$ is:

$$\vec{w} \cdot \vec{x} \simeq \rho \hat{w} \cdot \vec{x} = \rho \sum_{j=0}^{N-1} \hat{w}_j x_j \quad (4)$$

Since each $\hat{w}_j$ is an integer, each product $\hat{w}_j x_j$ can be expressed as:

$$\hat{w}_j x_i = \begin{cases} 0 & \text{if } \hat{w}_j = 0 \\ \underbrace{x_j + \ldots + x_j + \ldots + x_j}_{|\hat{w}_j| \text{ times}} & \text{if } \hat{w}_j > 0 \\ \underbrace{-x_j + \ldots - x_j + \ldots - x_j}_{|\hat{w}_j| \text{ times}} & \text{if } \hat{w}_j < 0 \end{cases} \quad (5)$$

Equation 5 seems quite obvious but, when considered together with (3), it follows that $\sum_{j=0}^{N-1} \hat{w}_j x_j$ can be calculated with exactly Q additions and/or subtractions and no multiplications, regardless of the dimensionality of the vectors involved. Scaling the result by ρ completes the dot product.

Therefore the dot product in (1) can be approximated using the properties of PVQed vectors knowing that, at most, Q additions will be required. This can be done by an accumulator (in Q cycles) using (5), without a multiplier. With Q/N=1-2, a good accuracy is maintained. If Q/N>1, Q>N (a MAC will take N cycles) and thus the performance of the PE will be reduced but with the benefit of a much smaller design (multipliers tend to be about one order of magnitude larger than adders).

The result only needs scaling by ρ at the end of dot product, possibly incorporated in other scaling operations (such as batch normalisation) or incorporated in the non-linearity calculation of a CNN.

Eq. (5) provides us with a practical way to calculate $\sum_{j=0}^{N-1} \hat{w}_j x_j$ using the accumulator in Fig.1b. In fact, if the sign of the weight $w_j$ is positive, we add $x_j$ to the accumulator $w_j$ times, otherwise we subtract it. The sign controls the add/sub behaviour of the accumulator. Weights can be run-length encoded and/or compressed as before. The result is guaranteed to take Q cycles because of (3).

## 3.1. PVQed Weight Statistics

The table below shows TinyYolo v3 weight statistics after PVQ. The first column indicates the layer number. Only convolutional layers are listed as they are relevant to this discussion. The kernel column shows the kernel for that layer. The Q/N ratio is 3/2 except for first layer that requires less quantization. Other columns show the distribution of the values of the weights.

| #  | Kernel      | Q/N | % 0   | % ±1  | % ±2-3 | % ±4-7 | % ±8-15 | % ±16-31 | % ±32-63 |
|----|-------------|-----|-------|-------|--------|--------|---------|----------|----------|
| 0  | 3x3x3x16    | 4   | 14.12 | 18.98 | 24.77  | 25.00  | 15.50   | 1.62     | 0.00     |
| 2  | 3x3x16x32   | 3/2 | 38.80 | 32.53 | 19.49  | 7.81   | 1.32    | 0.043    | 0.00     |
| 4  | 3x3x32x64   | 3/2 | 34.96 | 34.32 | 22.38  | 6.99   | 1.22    | 0.12     | 0.0054   |
| 6  | 3x3x64x128  | 3/2 | 29.69 | 37.16 | 24.71  | 7.55   | 0.85    | 0.041    | 0.0014   |
| 8  | 3x3x128x256 | 3/2 | 27.64 | 36.48 | 27.35  | 7.60   | 0.86    | 0.052    | 0.0041   |
| 10 | 3x3x256x512 | 3/2 | 25.63 | 37.39 | 29.06  | 7.29   | 0.60    | 0.028    | 0.0019   |
| 12 | 3x3x512x1024| 3/2 | 1.07  | 72.52 | 21.90  | 4.26   | 0.23    | 0.011    | 0.0004   |
| 13 | 1x1x1024x256| 3/2 | 25.84 | 35.97 | 30.30  | 7.23   | 0.62    | 0.036    | 0.0004   |
| 14 | 3x3x256x512 | 3/2 | 27.58 | 36.32 | 26.58  | 8.75   | 0.77    | 0.001    | 0.0001   |
| 15 | 1x1x512x255 | 3/2 | 34.22 | 29.50 | 24.07  | 11.54  | 0.65    | 0.016    | 0.0008   |
| 18 | 1x1x256x128 | 3/2 | 24.08 | 35.79 | 32.41  | 7.53   | 0.19    | 0.000    | 0.00     |
| 20 | 3x3x384x256 | 3/2 | 28.43 | 34.67 | 27.65  | 8.28   | 0.95    | 0.0028   | 0.0001   |
| 21 | 1x1x256x255 | 3/2 | 33.56 | 31.72 | 22.06  | 12.06  | 0.59    | 0.0031   | 0.00     |

Table 1: TinyYolo v3 weight statistics after PVQ.

It is worth noting that for all the layers (except for the first that requires less quantization), the magnitude of 90% of the weights is less than 4. This has also means that PVQed weights are highly compressible, as shown in the table below.

| # | Kernel | Bits | Bits/weigth |
|---|---|---|---|
| 0 | 3x3x3x16 | 1,913 | 4.43 |
| 2 | 3x3x16x32 | 13,061 | 2.94 |
| 4 | 3x3x32x64 | 54,165 | 2.94 |
| 6 | 3x3x64x128 | 223,095 | 3.03 |
| 8 | 3x3x128x256 | 910,072 | 3.09 |
| 10 | 3x3x256x512 | 3,587,849 | 3.04 |
| 12 | 3x3x512x1024 | 11,101,832 | 2.35 |
| 13 | 1x1x1024x256 | 812,074 | 3.10 |
| 14 | 3x3x256x512 | 3,622,674 | 3.07 |
| 15 | 1x1x512x255 | 410,331 | 3.14 |
| 18 | 1x1x256x128 | 100,992 | 3.08 |
| 20 | 3x3x384x256 | 2,702,768 | 3.05 |
| 21 | 1x1x256x255 | 203,784 | 3.12 |
|  | Total | 23,744,611 | 2.68 |

Table 2: TinyYolo v3 PVQed weight compressed as run-lengths and entropy coded.

The compressed size is estimated from the well known relationship -log$_2$(P) where P is the probability of the symbol being encoded. This establishes a lower bound for compression but it is also achievable with arithmetic coding. Because the weights do not change during inference, the underlining probability model is statically determined and the symbol probabilities are exact.

Two examples of video sequences processed by Tiny Yolo v3 (before and after PVQ) can be found on Youtube [6].

## 4. Bit Layer MAC

The accumulator in Fig.1b takes advantage of the properties of dot products with PVQed vectors to eliminate multipliers. Unfortunately doing multiplications by repeated additions is certainly not the most efficient way.

On the other hand, using MACs like in Fig.1a is certainly faster but it's not very efficient either. In fact, if we look at Tab.1, we notice that a large percentage of the weights after PVQ are very small. However, the multiplier in the MAC must be sized to cater for the largest expected weight (7 bits including the sign, but it could be even larger in other cases). This means that most of the times the multiplier designed for 7+ bits numbers will only be multiplying by 1-2 bit numbers. The extra logic is not required most of the times but it will still be using power and area.

This section will be addressing these issues.

In order to simplify the discussion, we will start with assuming the weights to be positive integers. Negative weights will be dealt with later. Let's consider again the dot product between a vector $\vec{x}$ and the weights vector $\hat{w}$. Each component $\hat{w}_j$ of $\hat{w}$ is a positive integer and it can represented as a binary number $\hat{w}_j = \sum_{i=0}^{N_b-1} d_{ij} 2^i$ with $d_{ij} \in \{0,1\}$ :

$$\hat{w} \cdot \vec{x} = \sum_{j=0}^{N-1} \hat{w}_j x_j = \sum_{j=0}^{N-1} \left( \sum_{i=0}^{N_b-1} d_{ij} 2^i \right) x_j = \sum_{i=0}^{N_b-1} \left( \sum_{j=0}^{N-1} d_{ij} x_j \right) 2^i =$$

$$\left( \ldots \left( \left( \sum_{j=0}^{N-1} d_{N_b-1\,j} x_j \right) 2 + \sum_{j=0}^{N-1} d_{N_b-2\,j} x_j \right) 2 + \ldots \right) 2 + \sum_{j=0}^{N-1} d_{0j} x_j \quad (6)$$

Where N is the dimensionality of the vectors and $N_b$ the number of bits of the weights. Eq. 6 shows that $\hat{w} \cdot \vec{x}$ can be calculated by starting with the contribution $\sum_{j=0}^{N-1} d_{N_b-1,j} x_j$ of all the most significant bits (MSB) of the weights. We then multiply by 2 and add the contribution of the second MSB. We continue to do that until we reach the least significant bit (LSB). If we look at the matrix $d_{ij}$ of binary digits we are performing multiply and accumulate operations from the MSB row to the LSB one, one bit layer at the time. Hence the name bit layer MAC (BLMAC). We will indicate non-zero elements in a bit layer ($d_{ij} \neq 0$) as "pulses".

Let's consider a simple example where $\hat{w} = (1, 27, 7, 0, 2)$ and $\vec{x} = (x_0, x_1, x_2, x_3, x_4)$, Nb=N=5. Then $\hat{w} \cdot \vec{x} = x_0 + 27 x_1 + 7 x_2 + 2 x_4$. We can calculate the same using equation 6.

These vectors are shown on Tab.3 with a column containing the binary representation of a weight.

| **4** | 0 | 1 | 0 | 0 | 0 |
|---|---|---|---|---|---|
| **3** | 0 | 1 | 0 | 0 | 0 |
| **2** | 0 | 0 | 1 | 0 | 0 |
| **1** | 0 | 1 | 1 | 0 | 1 |
| **0** | 1 | 1 | 1 | 0 | 0 |
|  | 1 | 27 | 7 | 0 | 2 |
|  | $x_0$ | $x_1$ | $x_2$ | $x_3$ | $x_4$ |

*Table 3: Bit layer MAC example.*

We start from row or layer 4, the MSB layer. With only one pulse, we get $x_1$. We now multiply by 2 and continue to add the contribution of layer 3: $2x_1+x_1 = 3x_1$. Again we multiply by 2 and add the contribution of layer 2: $6x_1+x_2$. We continue with layer 1: $2(6x_1+x_2)+x_1+x_2+x_4 = 13x_1+3x_2+x_4$. Finally, with layer 0: $2(13x_1+3x_2+x_4)+x_0+x_1+x_2 = x_0+27x_1+7x_2+2x_4$. Which is the expected result.

Fig.1c shows one of the possible architecture for a BLMAC. The S input controls the add/subtract behaviour of the accumulator. Since this is only relevant to negative weights, we will disregard this input for now and only assume addition capabilities. Multiplying a value by 2 in binary is a simple shift to the left, indicated by <<1.

Starting from the MSB layer $i=N_b-1$, we present at the input all the values xj for which dij≠0 which will be added to the accumulator every clock cycle. Once all the pulses in a layer are exhausted the `shift` input will cause the accumulator to shift one position to the left. We then continue to process all the pulses in the next layer and so on, until all the layers are processed. We note that :

- A BLMAC is essentially an add/sub accumulator plus a 2:1 multiplexer to select the shift operation.
- For CNNs, the dimensionality N of a dot product is much larger than the number of bits $N_b$ of a weight (a few thousands vs. no more than a dozen). Therefore the $N_b$ shifts required by the BLMAC can be considered negligible. Besides, the architectures in Fig.1c could be modified to perform the shift in the same clock cycle.
- If we skip all the zero $d_{ij}$ values, then the number of additions needed to calculate $\hat{w} \cdot \vec{x}$ is equal to the total number of pulses in all the weights. This is analogous to the sparse weight situation previously discussed, except that the "weights" are now either 0 or 1 (0 and +/-1 with negative numbers).
- The smaller the number of pulses, the faster is to calculate $\hat{w} \cdot \vec{x}$. This architecture exploits sparsity at bit level, not just at weight level.

- A BLMAC is naturally a variable precision MAC: it can deal with different size weights efficiently, without the need of sizing the design according to the worst case (largest) weight. The same architecture will work with binary, integer and even floating point weights. The latter might appear surprising until one considers that a fp32 number can be seen as a very large but bit sparse integer (up to a scaling factor) with, at most, 24 non zero bits.
- The design in Fig.1c is based on Eq. 6. It starts from the MSB bit layer and uses left shifts. The same result is obtained by starting from the LSB and using right shifts. With each right shift one bit of the final result is output. This version can have some advantages over the other, such as smaller adder and accumulator as the bits that are right shifted out are no longer part of the computation.

We now come to negative weights. The simplest way is to encode a weight as two's complement number. The same design in Fig.1c can be used. Incoming $x_j$ values will be subtracted from the accumulator when processing the sign layer (with S input = 1). For all the other bit layers, $x_j$ values will be added (as before).

This is because, in two's complement representation, the sign bit is a negative power of two. For example, if we have a 3 bit numbers and we want add a sign to it, the value -1 will be $1111_2 = -2^3+2^2+2^1+2^0$.

There is a problem with this representation, however. It requires a lot of pulses (think $-1=1111_2$ vs. $1=0001_2$) for very common weights such -1, or -2 (see Tab.1). This will have a negative impact on the processing speed as more pulses require more clock cycles.

This can be solved if we pass to a ternary representation with $d_{ij} \in \{-1,0,1\}$ but still represent weights as

$$\hat{w}_j = \sum_{i=0}^{N_b-1} d_{ij} 2^i$$

. Now positive and negative numbers use exactly the same number of pulses. For example, if we look at number 5, $101_2$ in binary, then -5 will be $(-1,0,-1)=-2^2-2^0$.

Moreover, we can now reduce the number of pulses in some cases. For example $11111_2$ can also be represented by $(1,0,0,0,0,-1)$: 2 pulses instead of 5. Another case is $11011_2 = (1,0,0,-1,0,-1)$: 3 pulses instead of 4. There are many other similar cases and this representation can help reduce the number of pulses and hence increase the speed. We can now repeat the same example of Tab. 3 with the new encoding.

| 5 | 0 | 1 | 0 | 0 | 0 |
|---|---|---|---|---|---|
| 4 | 0 | 0 | 0 | 0 | 0 |
| 3 | 0 | 0 | 1 | 0 | 0 |
| 2 | 0 | -1 | 0 | 0 | 0 |
| 1 | 0 | 0 | 0 | 0 | 1 |
| 0 | 1 | -1 | -1 | 0 | 0 |
|   | 1 | 27 | 7 | 0 | 2 |
|   | $x_0$ | $x_1$ | $x_2$ | $x_3$ | $x_4$ |

Table 4: Same BLMAC example with new encoding.

Starting from row 5 and get $x_1$. Next row there are no pulses, so we just shift : $2x_1$. At row 3 we shift again and add $x_2$, obtaining $4x_1+x_2$. At row 2 we shift and subtract x1 : $8x_1+2x_2-x_1 = 7x_1+2x_2$. At row 1 we shift and add x4 : $14x_1+4x_2+x_4$. Finally comes row 0, we shift and add/subtract the contributions according to the weights: $2(14x_1+4x_2+x_4)+x_0-x_1-x_2 = x_0+28x_1-x_1+8x_2-x_2+2x_4 = x_0+27x_1+7x_2+2x_4$.

The design of Fig.1c can still be used. This time, when a pulse is positive, $x_j$ values will be added, subtracted otherwise. The polarity of the pulse will control the S input.

It is also worthwhile to show how the same result is also obtained by starting from the LSB layer and shift right at the end of each bit layer. Every time this architecture shifts the accumulator right, a bit of the result is output. Here, in order to show the full result, all the inputs will be scaled by $32=2^5$ (5 is the number of bit layers).

So, starting from the LSB layer 0 : $32x_0-32x_1-32x_2$. Next layer 1, right shift : $16x_0-16x_1-16x_2$ and add $32x_4$. Result is $16x_0-16x_1-16x_2+32x_4$. Layer 2, shift and subtract $32x_1$ : $8x_0-8x_1-8x_2+16x_4-32x_1 = 8x_0-40x_1-8x_2+16x_4$. Layer 3, shift and add $32x_2$ : $4x_0-20x_1-4x_2+8x_4+32x_2 = 4x_0-20x_1+28x_2+8x_4$. Layer 4 is just a shift : $2x_0-10x_1+14x_2+4x_4$. Finally layer 5, shift and add $32x_1$ : $x_0-5x_1+7x_2+2x_4+32x_1 = x_0+27x_1+7x_2+2x_4$.

One final note to stress the differences and the advantages of a BLMAC over a serial MAC (i.e. a MAC that performs the multiplication serially and then adds the result to an accumulator (Fig.1d)).

In a serial MAC, two registers are required: one for the multiplication by shifts and adds and one to add the result to the accumulator. Two adders are also needed, although a single one could potentially be shared. This is about twice the size of a BLMAC.

Another problem is due to the shifts. For example, if we serially multiply $x_j$ by $10001_2$, we start with $x_j$, we shift it 4 times to the left and then add $x_j$ again. If these shifts are performed serially, this will take 5 cycles versus 2 in a BLMAC. This is because a BLMAC only perform a shift at the end of each bit layer instead of every single weight. Adding a barrel shifter could allow multiple shifts in the same clock cycle. This again adds complexity. Moreover, when adding a barrel shifter one would need to decide what is the maximum number of simultaneous shifts. More shift means larger logic, mostly underutilised with infrequent large weights. Few shifts means additional cycles and lower performance in some cases. This not a problem for BLMAC that handle variable precision naturally without costly barrel shifters.

In conclusion, a BLMAC is substantially different, smaller and faster than a serial MAC.

## 4.1. BLMAC Performance

The table below shows, for integers up to 24 bits, the average and the maximum number of pulses. Column 7, for example, contains these statistics for all the 7 bit integers (i.e. all the integers between 0 and +127 included).

| $N_b$ | 1 | 2 | 3 | 4 | 5 | 6 | 7 | 8 | 9 | 10 | 11 | 12 | 13 | 14 | 15 | 16 | 17 | 18 | 19 | 20 | 21 | 22 | 23 | 24 |
|---|---|---|---|---|---|---|---|---|---|---|---|---|---|---|---|---|---|---|---|---|---|---|---|---|
| Avg | 0.5 | 1.0 | 1.37 | 1.75 | 2.09 | 2.44 | 2.77 | 3.11 | 3.44 | 3.77 | 4.11 | 4.44 | 4.78 | 5.11 | 5.44 | 5.77 | 6.11 | 6.44 | 6.78 | 7.11 | 7.44 | .7.78 | 8.11 | 8.44 |
| Max | 1 | 2 | 2 | 3 | 3 | 4 | 4 | 5 | 5 | 6 | 6 | 7 | 7 | 8 | 8 | 9 | 9 | 10 | 10 | 11 | 11 | 12 | 12 | 13 |

Table 5: Average and maximum number of pulses for numbers of given bit size.

So column 7 tells us that any integer in such range can be encoded with a maximum of 4 pulses. This is also the maximum number of cycles that a BLMAC will take to multiply by a weight in the same range and, if the weights are uniformly distributed, we can expect to take an average of ~2.77 cycles each.

A negative number takes exactly the same number of cycles, as the only difference is the inverted pulse sign. Thus column 7 also includes the case of 8 bits signed numbers.

In CNNs, the weights are far from uniformly distributed and often exhibit a Laplacian/Gaussian distribution as shown in Tab. 1 for TinyYolo v3 and so a BLMAC will be faster than the expected average of Tab.5.

## 4.2. PVQed Weights Compressed as Bit Layers

All the weights encoding and compression considerations previously described are still valid. The main difference is that now, for the pairs (ZRUN,W), W can only be ±1. Also, EOR now encodes the end of a bit layer. We can see how run-length encoding works in practice on the example in Tab.4.

Starting from layer 5, we can scan from left to right. The direction or, indeed, the order in which the pulses are scanned within a layer is not important as long as we are consistent during run-length expansion.

The first pulse is a 1, so W=1. Between the start and this pulse there is one zero, so ZRUN=1. The first run-lenght will then be (1,1). After that, there are no more pulses in this row. We can skip to the end with an EOR. Row 4 has no pulses, so it's just an EOR. Row 2 has a pulse two zeros from the start : (2,1) plus an EOR. Row 1 is encoded as (4,1). Now, because the last position in the row is a pulse, an EOR is implicit and it can be inferred during run-length expansion. Row 0 is formed by three consecutive pulses that are encoded as (0,1),(0,-1) and (0,-1). An EOR completes the run-length encoding process.

The table below shows an example of the PVQed Tiny Yolo v3 weights compressed as bit layers run-lengths.

| # | Kernel | Bits | Bits/weigth |
|---|---|---|---|
| 0 | 3x3x3x16 | 2,309 | 5.35 |
| 2 | 3x3x16x32 | 14,145 | 3.07 |
| 4 | 3x3x32x64 | 57,710 | 3.13 |
| 6 | 3x3x64x128 | 238,560 | 3.24 |
| 8 | 3x3x128x256 | 984,092 | 3.34 |
| 10 | 3x3x256x512 | 3,893,419 | 3.30 |
| 12 | 3x3x512x1024 | 13,478,875 | 2.86 |
| 13 | 1x1x1024x256 | 871,844 | 3.26 |
| 14 | 3x3x256x512 | 3,799,890 | 3.22 |
| 15 | 1x1x512x255 | 437,565 | 3.35 |
| 18 | 1x1x256x128 | 111,388 | 3.40 |
| 20 | 3x3x384x256 | 2,867,268 | 3.25 |
| 21 | 1x1x256x255 | 219,497 | 3.36 |
| Total | | 26,976,564 | 3.05 |

Table 6: TinyYolo v3 PVQed weight compressed as bit layers.

The methodology is the same as in the previous example. Even though the data is the same, compression along bit layers is ~13.5% lower. This is due to the fact that, in the previous case, the sign bit is encoded only once per weight whereas, in this case, every pulse carries a sign.There are more complex schemes that get around this but they are less practical for hardware implementation.

# 5. Performance Comparison

This section compares the performance of architectures in Fig.1 in the case of the CNN TinyYolo v3. Tab.7 shows some statistics for said CNN after PVQ.

| # | Kernel | N | Q | NZ | $N_3$ |
|---|---|---|---|---|---|
| 0 | 3x3x3x16 | 432 | 1,728 | 371 | 558 |
| 2 | 3x3x16x32 | 4,608 | 6,912 | 2,820 | 3,375 |
| 4 | 3x3x32x64 | 18,432 | 27,648 | 11,989 | 14,180 |
| 6 | 3x3x64x128 | 73,628 | 110,582 | 51,839 | 60,814 |
| 8 | 3x3x128x256 | 294,912 | 442,368 | 213,392 | 251,784 |
| 10 | 3x3x256x512 | 1,179648 | 1,769,472 | 877,721 | 1,033,521 |
| 12 | 3x3x512x1024 | 4,718,592 | 7,077,888 | 4,668,036 | 5,221,710 |
| 13 | 1x1x1024x256 | 262,144 | 393,216 | 194,402 | 229,973 |
| 14 | 3x3x256x512 | 1,179,648 | 1,769,472 | 854,302 | 1,020,873 |
| 15 | 1x1x512x255 | 130,560 | 195,840 | 85,879 | 106,308 |
| 18 | 1x1x256x128 | 32,768 | 49,152 | 24,299 | 29,299 |
| 20 | 3x3x384x256 | 884,736 | 1,327,104 | 633,226 | 752,090 |
| 21 | 1x1x256x255 | 65,280 | 97,920 | 43,369 | 53,112 |
| Total | | 2,140,369,920 | 3,354,324,480 | 1,685,206,900 | 1,974,123,320 |

Table 7: Architectures performance figures for TinyYolo v3 CNN PVQed weights.

Column N shows the number of weights in the kernel (also the number of dimensions when the kernel is flattened into a vector). For example, for layer 4, the kernel is 3x3x32x64 = 18,432 elements. This is also the number of cycles taken by the MAC in Fig.1a (assuming a single cycle for the multiplication and the addition).

Column Q contains the PVQ Q parameter. The Q/N ratio is 3/2 except for first layer that requires less quantization. This is also the number of cycles taken by the accumulator in Fig.1b. For example, for the same layer 4, it will only require Q= 27,648 additions.

Column NZ shows the total number of non-zero weights after PVQ for a given layer. This is also the number of cycles taken by the MAC in Fig.1a when skipping all the zero weights.

Column $N_3$ shows the number of cycles taken by the BLMAC in Fig.1c for a given layer.

The total indicates the number of cycles for a 416x320 input image.

A MAC skipping zero weights is clearly the fastest (NZ column).

This is not the whole story, however. A MAC, especially a single cycle one, requires at least one order of magnitude more silicon resources than a simple accumulator. This is amplified by the potentially large number of MACs substituted (hundreds working in parallel). The simpler, shallower logic of Fig.1b is also likely to run faster.

The advantage is even bigger for BLMACs : only ~17% slower than a MAC skipping zero weights and ~1.7x faster than the simple accumulator with minimal additional logic.

Finally, a BLMAC is slightly faster than a naive MAC architecture that does not skip zero weights. In this case, a MAC (one multiplication and one addition) per weight is substituted by an average of less than one addition in BLMACs. S

Since the original Tiny Yolo v3 network implementation is in floating point, this means that the combination of PVQ and architecture III allows to substitute one floating point MAC operation per weight with less than one integer addition.

The totals indicates the number of cycles for a complete 416 x 320 input image. Note that this is not simply the sum of the columns above but takes into account the size of the feature maps of each layer.

Based on these numbers, it takes ~1.5 integer additions/weight for PVQ (unsurprisingly, since the N/Q=3/2 for most of the layers) and ~0.92 integer additions/weight for PVQ+BLMAC.

# 6. Another Example: FIR filters

FIR filters are another potential application of the techniques described in this paper.

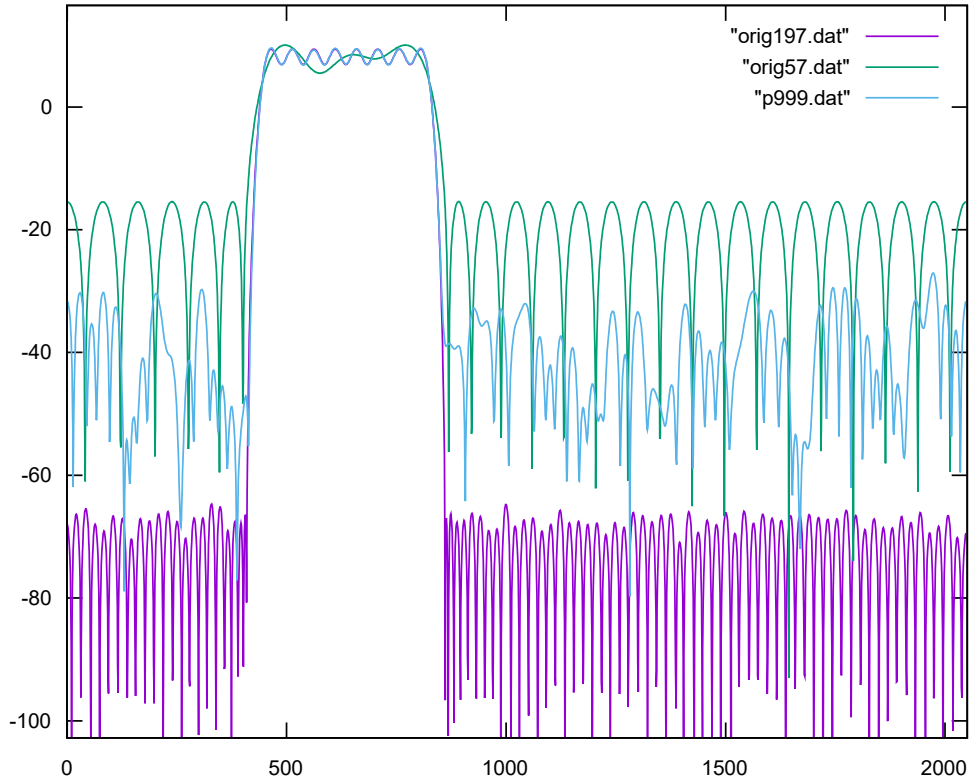

Figure 2: FIR filters frequency response.

Fig.2 shows the frequency response of a linear phase FIR bandpass filter with 197 taps (**orig197**). The filter pass band is between 220 and 400 Hz, with sharp attenuation to below -60 dB. This filter was created the free online tool Tfilter [7] with 16 bits integers weights.

Fig.2 shows the frequency response of the same filter with weights PVQed with Q=999 (**p999**). A 57 tap FIR filter was also created for comparison and its frequency response is shown in Fig.2 (**orig57**). If we consider that the computational cost of a 16 bit multiplication is ~16x the one of a single addition, then **orig57** and **p999** have a similar cost but the latter offers a better attenuation.

| Operations | MAC | BLMAC | PVQ | PVQ+BLMAC |
|---|---|---|---|---|
| Additions | 197 | 542 | 999 | 238 |
| Multiplications | 197 | 0 | 1 | 1 |

Table 8 FIR filter **orig197** comparison.

The MAC column in Tab.8 shows the number of additions and multiplications required by **orig197**. The number of cycles will also be 197 when calculated with a MAC (Fig.1a).

The BLMAC column shows the number of additions (and cycles) required by a BLMAC (Fig.1c). Note that this is not an approximation. The weights are not quantized and the BLMAC result will be identical to the one of the MAC. The BLMAC is ~2.7x slower but is also ~16x smaller than a MAC. It is however twice as fast as it would be expected for 16 bits integers if they were uniformly distributed (column 15, Tab.5). This is another example demonstrating the effectiveness of a BLMAC at exploiting the weights' distribution, even in their original, unquantized form.

The PVQ column shows the number of additions (and cycles) required by an accumulator (Fig.1b) with PVQed weights (**p999**). A BLMAC, when working with the same PVQed weights (PVQ+BLMAC column) is more than 4x faster than a simple accumulator even both require a similar amount of resources. Note that the results with PVQed weights need to be scaled by ρ (see eq. 4).

One final observation : calculating the unquantized version of this filter **orig197** with a BLMAC is nearly twice as fast as calculating the PVQed version **p999**.

# 7. Conclusion

This paper has introduced different architectures for dot product calculation working in conjunction with PVQ. Some characteristics include:

- Efficient support for sparse weights and their compression
- Elimination of multipliers
- Support for variable precision
- Exploitation of weight sparsity at bit level (BLMAC)
- Low resources usage and shallow, fast logic

These architectures represent the building blocks for efficient CNN inference processors in a low power and high speed environment.